\documentclass[letterpaper]{article} % DO NOT CHANGE THIS
\usepackage{aaai23}  % DO NOT CHANGE THIS
\usepackage{times}  % DO NOT CHANGE THIS
\usepackage{helvet}  % DO NOT CHANGE THIS
\usepackage{courier}  % DO NOT CHANGE THIS
\usepackage[hyphens]{url}  % DO NOT CHANGE THIS
\usepackage{graphicx} % DO NOT CHANGE THIS
\usepackage{natbib}  % DO NOT CHANGE THIS AND DO NOT ADD ANY OPTIONS TO IT
\usepackage{caption} % DO NOT CHANGE THIS AND DO NOT ADD ANY OPTIONS TO IT
\usepackage{algorithm}
\usepackage{algorithmic}

\usepackage{newfloat}
\usepackage{listings}
\DeclareCaptionStyle{ruled}{labelfont=normalfont,labelsep=colon,strut=off} % DO NOT CHANGE THIS
\floatstyle{ruled}
\newfloat{listing}{tb}{lst}{}
\floatname{listing}{Listing}
\usepackage{subfigure}
\usepackage{amsmath}
\usepackage{amssymb}
\usepackage{color}

\title{Mind the Gap: Polishing Pseudo Labels for Accurate Semi-Supervised Object Detection}
\author{
    Lei Zhang\textsuperscript{\rm 1}\equalcontrib, 
    Yuxuan Sun\textsuperscript{\rm 1}\equalcontrib,
    Wei Wei\textsuperscript{\rm 1,}\textsuperscript{\rm 2}\thanks{Corresponding author}
}
\affiliations{
    \textsuperscript{\rm 1} School of Computer Science, Northwestern Polytechnical University, China \\
    \textsuperscript{\rm 2} Research \& Development Institute of Northwestern Polytechnical University in Shenzhen, China \\
    \{nwpuzhanglei,weiweinwpu\}@nwpu.edu.cn, sunyuxuan@mail.nwpu.edu.cn
}

\usepackage{bibentry}
\begin{document}

\maketitle

\begin{abstract}
    Exploiting pseudo labels (e.g., categories and bounding boxes) of unannotated objects produced by a teacher detector have underpinned much of recent progress in semi-supervised object detection (SSOD). However, due to the limited generalization capacity of the teacher detector caused by the scarce annotations, the produced pseudo labels often deviate from ground truth, especially those with relatively low classification confidences, thus limiting the generalization performance of SSOD. To mitigate this problem, we propose a dual pseudo-label polishing framework for SSOD. Instead of directly exploiting the pseudo labels produced by the teacher detector, we take the first attempt at reducing their deviation from ground truth using dual polishing learning, where two differently structured polishing networks are elaborately developed and trained using synthesized paired pseudo labels and the corresponding ground truth for categories and bounding boxes on the given annotated objects, respectively. By doing this, both polishing networks can infer more accurate pseudo labels for unannotated objects through sufficiently exploiting their context knowledge based on the initially produced pseudo labels, and thus improve the generalization performance of SSOD. Moreover, such a scheme can be seamlessly plugged into the existing SSOD framework for joint end-to-end learning. In addition, we propose to disentangle the polished pseudo categories and bounding boxes of unannotated objects for separate category classification and bounding box regression in SSOD, which enables introducing more unannotated objects during model training and thus further improves the performance. Experiments on both PASCAL VOC and MS-COCO benchmarks demonstrate the superiority of the proposed method over existing state-of-the-art baselines. The code can be found at https://github.com/snowdusky/DualPolishLearning.
\end{abstract}

\begin{figure}[t!]
\centering
\subfigure[]{
    \includegraphics[width=1.5in]{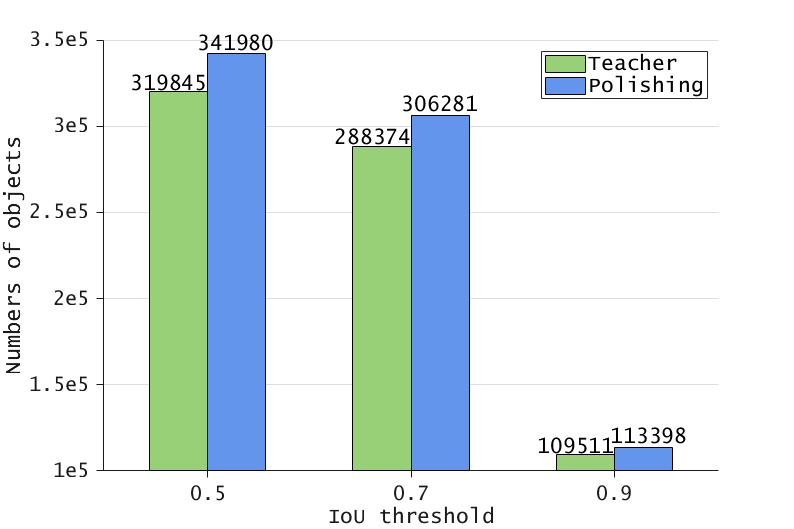}
}
\subfigure[]{
    \includegraphics[width=1.6in]{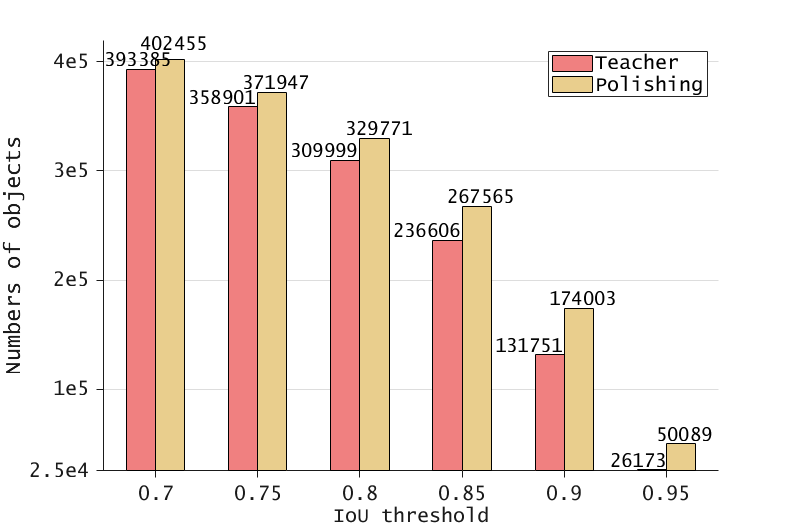}
}
% \vspace{-0.1cm}
% \setlength{\abovecaptionskip}{-0.5pt}
\caption{Quality of pseudo labels for unannotated objects produced by a teacher detector and polished by the proposed method on the MS-COCO dataset with $10\%$ labeled images. (a) The amount of unannotated objects with correct pseudo category labels, when IoU between the pseudo bounding boxes and the ground truth is larger than different thresholds. A larger amount indicates more accurate pseudo category labels. (b) The amount of unannotated objects with different quality of pseudo bounding boxes. The quality is measured by the IoU between the pseudo bounding box and ground truth. A larger amount indicates more accurate pseudo bounding boxes. As can be seen, 
% the amount of objects with high-quality pseudo labels is obviously increased.
through polishing the pseudo labels using the proposed method, the amount of objects with high-quality pseudo labels including both categories and bounding boxes is obviously increased.
}
\label{fig:coco_reg_polish}
% \vspace{-0.8cm}
\end{figure}

\section{Introduction}

Deep neural networks (DNNs) have achieved impressive progress in object detection, when being supervisedly trained with a large amount of annotated objects. However, it is often infeasible to collect such sufficient annotated objects in real applications, since the manual annotation is expensive and time-consuming. As a result, increasing efforts~\cite{zhou2021instant,tang2021humble,liu2021unbiased,xu2021end} commence investigating semi-supervised object detection (SSOD), which aims to achieve good generalization performance using only a few annotated objects together with extensive unannotated objects. Since DNNs with a few annotated objects are prone to be over-fitting, the key for SSOD is how to exploit the beneficial knowledge from extensive unannotated objects to regularize the supervised training of DNNs. To achieve this goal, a promising solution is the pseudo labeling technique~\cite{zoph2020rethinking} which aims at introducing unannotated objects with pseudo labels produced by a teacher detector to augment the scarce annotated objects for model training. %In general, at the beginning the teacher detector is supervisedly trained on the given annotated objects, and then it will be constantly updated based on the result of semi-supervised learning %using increasing unannotated objects with inferred pseudo labels together with the given annotated objects, until convergence~\cite{zhou2021instant,xu2021end}. 
Till now, many pseudo labeling based SSOD methods~\cite{sohn2020simple,zhou2021instant,tang2021humble,liu2021unbiased,xu2021end} have been proposed, which mainly focus on constructing various semi-supervised learning framework to exploit the pseudo labels of unannotated objects produced by the teacher detector for better generalization performance. Although these methods have achieved obvious progress in SSOD, due to the limited generalization capacity of the teacher detector caused by the scarce annotated objects, the pseudo labels produced for unannotated objects often deviate from ground truth, especially those with relatively low classification confidence, as shown in Fig.~\ref{fig:coco_reg_polish}. To mitigate the negative effect of inaccurate pseudo labels on SSOD, most existing methods resort to only selecting a limited amount of unannotated objects with extremely high classification confidences (e.g., \textgreater0.9), and thus fail to sufficiently exploit the beneficial knowledge in extensive unannotated objects as well as achieve pleasing generalization performance. Few attention has been paid to directly 
% resolve the inherent problem, viz., 
reduce the deviation between pseudo labels and ground truth and introduce more unannotated objects with high-quality pseudo labels for SSOD.

To fill this gap, we propose a dual pseudo-label polishing framework, which aims at learning to reduce the deviation between the pseudo labels produced by a teacher detector and the ground truth for unannotated objects to improve the generalization performance of SSOD. To achieve this goal, we first elaborately develop two differently structured polishing networks to separately polish the pseudo labels of categories and bounding boxes for unannotated objects. In particular, a multiple ROI features aggregation module is developed to embed the context knowledge of unannotated objects 
% into the network 
for pseudo bounding boxes polishing. Then, based on the given annotated objects, we employ a Gaussian random sampling strategy to synthesize paired pseudo labels 
% produced by the teacher detector 
and the corresponding ground truth, and utilize them to supervisedly train both polishing networks. By doing this, both polishing networks can learn to infer more accurate pseudo labels for unannotated objects and thus improve the generalization performance of SSOD. Moreover, such a dual polishing learning scheme can be seamlessly plugged into the existing SSOD framework~\cite{xu2021end} for joint end-to-end learning. In addition, we propose to disentangle the polished pseudo categories and bounding boxes of unannotated objects in SSOD, i.e., the polished pseudo category or bounding box for a given unannotated object is separately utilized for category classification or bounding box regression training in SSOD. This enables introducing more unannotated objects for SSOD and further improves the generalization performance. Experiments on both PASCAL VOC and MS-COCO benchmarks demonstrate the effectiveness of the proposed method in coping with SSOD.

In summary, this study mainly contributes in four aspects:
\begin{enumerate}
    \item We propose a dual pseudo-label polishing framework, which, to the best of our knowledge, takes the first attempt at learning to reduce the deviation between the pseudo labels and ground truth of unannotated objects in SSOD.
    \item We develop two different structured polishing networks and a dual polishing learning scheme, which enables data-driven learning to separately increase the accuracy of pseudo categories and bounding boxes of unannotated objects in an end-to-end SSOD framework.
    \item We propose to disentangle the polished pseudo categories and bounding boxes of unannotated objects to introduce more unannotated objects for SSOD. 
    \item We demonstrate new state-of-the-art SSOD performance on both PASCAL VOC and MS-COCO benchmarks. 
\end{enumerate}

\section{Related Work}
\subsection{Semi-supervised Learning}
Semi-supervised learning (SSL) aims to utilize extensive unannotated data to mitigate the over-fitting problem caused by training complicated model using scarce annotated data. Most existing SSL methods are proposed for image classification tasks, which can be roughly categorized into two groups, including consistency constrained methods~\cite{berthelot2019mixmatch,miyato2018virtual,xie2020unsupervised,sajjadi2016regularization,tarvainen2017mean,berthelot2019remixmatch} and pseudo-labeling based methods~\cite{lee2013pseudo,xie2020self,zhai2019s4l,bachman2014learning,arazo2020pseudo,iscen2019label}. The consistency constrained methods utilize a smoothness assumption and encourage the model prediction invariant on differently augmented views of the same image. For example, Xie et al.~\cite{xie2020unsupervised} %and Berthelot et al.~\cite{berthelot2019remixmatch} 
demonstrate that data-augmentation schemes, e.g., CutOut~\cite{devries2017improved}, RandAugment~\cite{cubuk2020randaugment}, 
% and CTAugment~\cite{berthelot2019remixmatch},
can be enormously helpful as strong augmentation for SSL. Pseudo-labeling based methods %also known as self-training, 
%aim to
turn to generate high-quality pseudo-labels for unannotated data using the model pre-trained on a few annotated data and retrain the model on both annotated data and unannotated data with pseudo labels. 
% For example, Iscen et al.~\cite{iscen2019label} employ a transductive label propagation method to generate pseudo-labels for the unlabeled data and train a deep neural network. 
Recently, Sohn et. al~\cite{sohn2020fixmatch} propose to 
integrate both ideas mentioned above and consequently lead to obvious performance improvement. In a specific, %while vastly simplify the overall method. In  
it firstly generates pseudo labels on weakly-augmented unlabeled data using high-confidence predictions and then utilizes them to supervise the training on strongly-augmented version of the same image. Although the proposed method follows the pseudo-labeling idea, it mainly focuses on reducing the bias between pseudo labels and the ground truth for better SSOD.

\begin{figure*}[h]
\centering
\includegraphics[width=5.4in, height=2.0in]{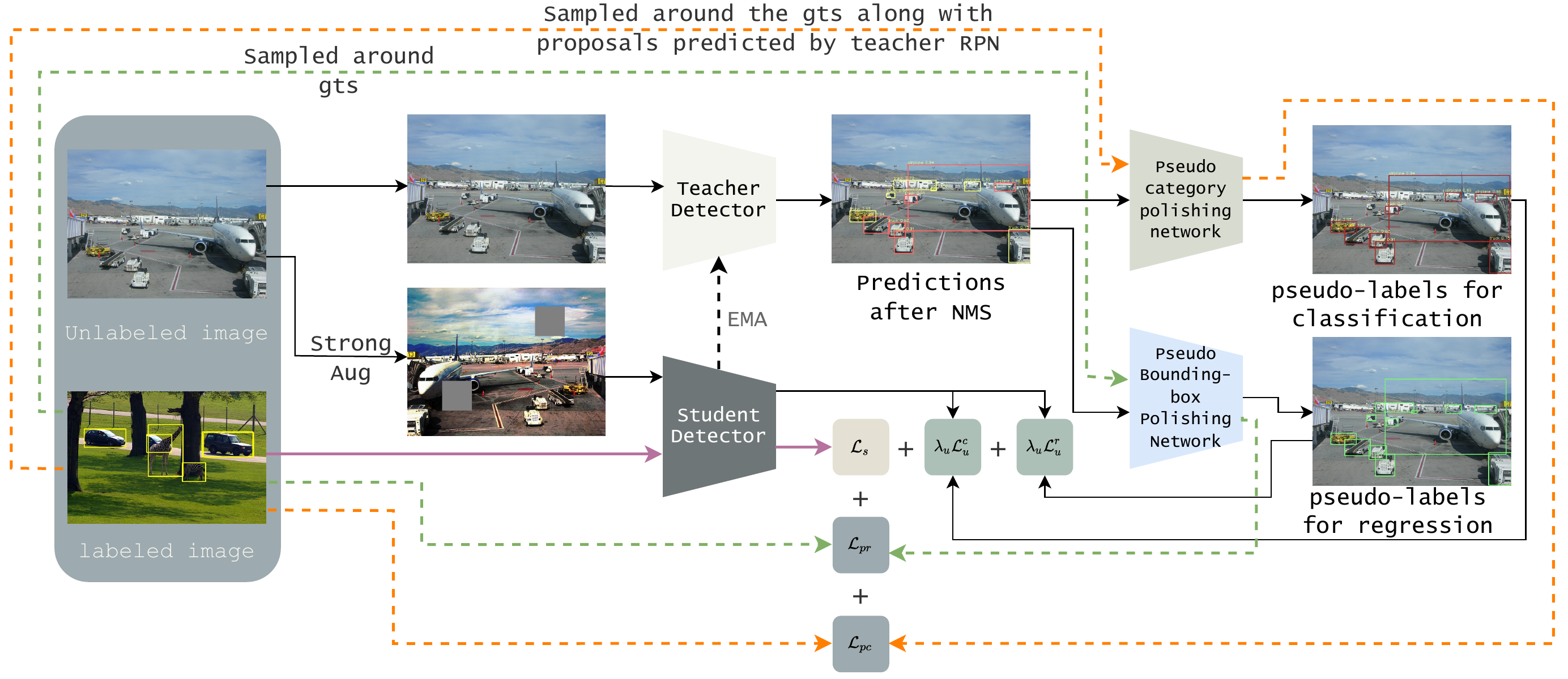}
\caption{Overview of the proposed dual pseudo-label polishing framework. 
We introduce two extra polishing networks to refine the initial pseudo labels produced by the teacher detector. 
In addition, the polished pseudo categories and bounding boxes are separately utilized to regularize the category classification and bounding-box regression heads of the student detector during training. The colored lines indicate the supervised training process of SSOD on annotated objects. The colored dotted lines sketch the dual polishing learning on the annotated data to optimize both polishing networks, while the black dotted line represents the process of EMA~\cite{tarvainen2017mean} which gradually updates the teacher detector based on the student one. $\mathcal{L}_u^c$ and $\mathcal{L}_u^r$ are the classification and regression parts of the pseudo supervised loss $\mathcal{L}_u$, while $\mathcal{L}_{pc}$ and $\mathcal{L}_{pr}$ are classification and regression parts of the loss $\mathcal{L}_p$ for dual polishing learning. Best view in color.
}
\label{fig:framework}
\end{figure*}

\subsection{Semi-supervised Object Detection}
Object detection~\cite{NIPS2015_14bfa6bb,Lin_2017_ICCV,Lin_2017_CVPR,redmon2018yolov3,zhang2020bridging} is a fundamental task in computer vision domain. Similar as semi-supervised image classification, SSOD aims to relieve the over-fitting problem caused by scarce object annotation using extensive unannotated objects. Till now, increasing effort, especially these pseudo labeling based methods~\cite{sohn2020simple,zhou2021instant,tang2021humble,liu2021unbiased,xu2021end}, have been made in SSOD. For example, inspired by the seminal SSL work~\cite{sohn2020simple} that introduces a weak-strong data augmentation scheme for SSL in classification, some works~\cite{zhou2021instant,tang2021humble,liu2021unbiased,xu2021end} integrate such a scheme with a mean teacher strategy~\cite{tarvainen2017mean} and establish a strong baseline for SSOD. Liu et al.~\cite{liu2021unbiased} turn to utilize some data augmentation schemes, e.g., MixUp~\cite{zhang2018mixup} and Mosaic~\cite{bochkovskiy2020yolov4} to introduce a number of reliable objects in the unannotated images based on data augmentation. Tang et al.~\cite{tang2021humble} utilize a light-weighted detection-specific data ensemble for base detector to generate more reliable pseudo-labels. 
Very recently, Xu et al.~\cite{xu2021end} develop an end-to-end SSOD framework, which can gradually produce pseudo labels for unannotated objects during the curriculum of updating the base detector, and thus achieve the state-of-the-art SSOD performance. While these methods have made obvious progress in SSOD, they still directly exploit the pseudo labels produced by the teacher detector. Thus, the inevitable deviation between pseudo labels and ground truth will cause the sub-optimal performance. In this study, we propose a pseudo-label polishing framework that learns to reduce the deviation of pseudo labels and ground truth using two well developed polishing networks. Although the recent work~\cite{zhou2021instant} also attempts to alleviate the pseudo-label deviation problem, it resorts to an unsupervised model ensemble scheme and thus fails to explicitly reduce the pseudo-label deviation as this study. 

\section{Methodology}
In this section, we first illustrate the proposed dual pseudo-label polishing framework. Then, we introduce two differently structured polishing networks and the dual polishing learning scheme, followed by the strategy of disentangling pseudo labels for SSOD.

\subsection{Dual Pseudo-Label Polishing Framework}
As shown in Fig~\ref{fig:framework}, the proposed dual pseudo-label polishing framework consists of three key modules, including a student detector that predicts the detection results on the test image, a teacher detector that produces the pseudo labels for the unannotated objects, and two developed polishing networks, i.e., a pseudo category polishing network and a pseudo bounding-box polishing network, that refine the produced pseudo labels through data-drivenly learning to reduce their deviation from ground truth using a dual polishing learning scheme.

In the proposed framework, we follow~\cite{xu2021end} to randomly initialize both the teacher and student detectors except the pre-trained feature extraction backbone. Then, the student detector is optimized on both the annotated objects (e.g., included in the labeled images) and the unannotated objects (e.g., included in the unlabeled images) with pseudo labels produced by the teacher detector, while the teacher detector is continuously updated by the weights of the student detector using the exponential moving average (EMA) strategy~\cite{tarvainen2017mean}. At the same time, two polishing networks are trained by a dual polishing learning scheme conducted on the labeled images with annotated objects. The details for both polishing networks and the dual polishing learning scheme will be given in the following subsections. Similar as~\cite{xu2021end}, the proposed framework can be trained in an end-to-end manner, and the overall training loss can be formulated as
\begin{equation}
    \label{eq:overall_loss}
    \mathcal{L} = \mathcal{L}_s + \lambda_u \mathcal{L}_u + \mathcal{L}_p,
\end{equation}

\begin{figure}[htbp]
\centering
\qquad\subfigure[Pseudo category polishing network.]{\includegraphics[width=0.8\linewidth]{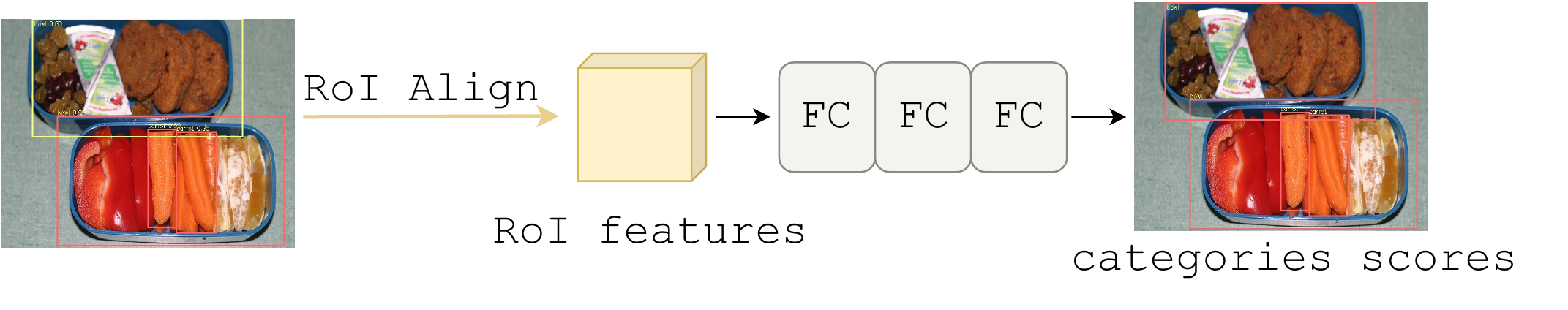}\label{fig:cls_polishing_net}}
\subfigure[Pseudo bounding-box polishing network.]{\includegraphics[width=0.85\linewidth]{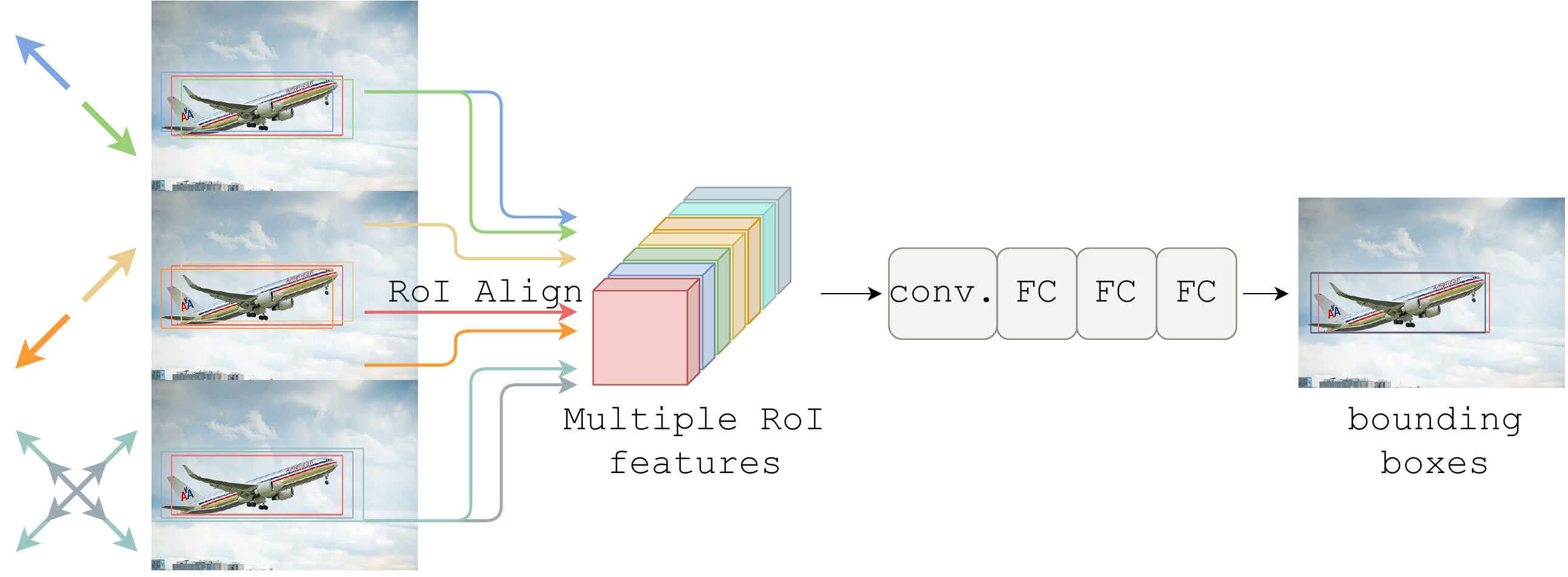}\label{fig:reg_polishing_net}}
\caption{Architecture of the proposed polishing networks} %where the BLOCK represents some FC layers. }% <------***** need to be modified>
\end{figure}

\subsection{Pseudo Category Polishing Network}
\label{sec::classification_polishing_network}
Considering that the deviation between the pseudo categories (e.g., coded in one-hot encoding scheme) and the corresponding ground truth is determined by the generalization performance of the teacher detector and has a continuous solution space, it is difficult to explicitly model such deviation using a regression model. To sidestep this problem, we propose to establish a pseudo category polishing network and employ it to re-predict the categories of the unannotated objects based on the pseudo bounding boxes produced by the teacher detector. Following this idea, we develop a three-layer network architecture as shown in Figure~\ref{fig:cls_polishing_net}. For a given unannotated object, the polishing network first extracts the ROI feature based on the pseudo bounding box and then transforms the flattened feature using two fully connected layers as well as finally outputs the refined category label. It is noticeable that the polishing network is different from the classification head in detector which predicts the category labels based on the proposal bounding boxes produced by the RPN module~\cite{NIPS2015_14bfa6bb,Lin_2017_ICCV}. More importantly, we will data-drivenly train the polishing network as Section~\ref{subsec:dual-pol}, which empowers the polishing network to learn to produce more accurate pseudo categories based on inaccurate bounding boxes, as shown in Figure~\ref{fig:coco_reg_polish}.

\subsection{Pseudo Bounding-Box Polishing Network}
\label{sec::regression_polishing_network}
Many previous literature~\cite{song2020revisiting} has proved that exploiting the context knowledge around objects is beneficial to accurately locate their bounding box. Inspired by this, we establish the pseudo bounding-box polishing network to appropriately mitigate the deviation between the pseudo bounding boxes produced by the teacher detector and the ground truth through sufficiently exploiting the context knowledge of the unannotated objects. Apparently, the key lies on sufficiently exploiting the context knowledge around the unannotated objects. Although the pseudo bounding boxes often show uncertainty, they have located the rough position of the unannotated objects, which can be utilized as clues to exploit the context knowledge. In other words, the context knowledge will be covered by some bounding boxes close to the pseudo one. Following this idea, we propose to augment the pseudo bounding boxes using shift and scaling schemes. In particular, without the loss of generality, we shift each pseudo bounding box along the four diagonal directions with a fixed distance, e.g., $\gamma \times d$ where $d$ denotes the diagonal length of the pseudo bounding box. In addition, we also enlarge the area of each pseudo bounding box using two fixed factors as $(1+2\times t \times\gamma)$ with $t\in\{1,2\}$. By doing these, we can augment each pseudo bounding box with 6 more context-related bounding boxes, as shown in Figure~\ref{fig:reg_polishing_net}. Then, we concatenate the ROI features extracted within each of them and utilize a four-layer network to predict the deviation between the pseudo bounding box and the ground truth for the considered unannotated objects, as shown in Figure~\ref{fig:reg_polishing_net}. Different from the bounding box regression head in detector, the polishing network can exploit the context knowledge to refine the pseudo bounding boxes produced by the teacher detector, and thus improves the accuracy of the pseudo bounding boxes for SSOD, as shown in Figure~\ref{fig:coco_reg_polish}. %In addition, it is noticeable that more refined augmentation can be utilized to more comprehensively model the context knowledge for better performance.

\subsection{Dual Polishing Learning}\label{subsec:dual-pol}
To appropriately reduce the deviation between pseudo labels produced by the teacher detector and the corresponding ground truth for unannotated objects, a promising way is to data-drivenly train both polishing networks developed in above. The key for this lies in collecting sufficient training pairs each of which consists of the pseudo labels 
% produced by the teacher detector 
and ground truth for unannotated objects and utilize them to train these two polishing networks in a supervised manner. Apparently, this is infeasible in the SSOD setting. Considering that only the scarce annotated objects have ground truth labels in SSOD, we present a Gaussian random sampling strategy to synthesize paired pseudo labels and ground truth using the annotated objects with their labels.   

Specifically, for the $i$-th annotated object with ground truth label ${\mathbf{g}}_i=\{c_i, \mathbf{x}_{ui}, \mathbf{x}_{di}\}$ where $c_i$ denotes the category label, while $\mathbf{x}_{ui}$ and $\mathbf{x}_{di}$ separately denote the upper-left and the lower-right coordination vectors of the bounding box, we synthesize the pseudo labels produced by the teacher detector through randomly sampling a large amount of pseudo bounding boxes, e.g., $\{\mathbf{x}^{pj}_{ui}, \mathbf{x}^{pj}_{di}\}$ denotes the $j$-th sampled bounding box, which can be formulated as  
\begin{equation}
\label{eq:random_sample}
\begin{aligned}
\mathbf{x}^{pj}_{ui} = \mathbf{x}_{ui} + \theta\times (\mathbf{s}_i \odot \mathbf{t}^j_{ui}); \\
\mathbf{x}^{pj}_{di} = \mathbf{x}_{di} + \theta\times(\mathbf{s}_i \odot \mathbf{t}^j_{di}); \\
\rm{s.t.} {\kern 4pt} \mathbf{t}^j_{ui},\mathbf{t}^j_{di} \sim \mathcal{N}(\mathbf{0}, \mathbf{I})
\end{aligned}
\end{equation}
where $\mathbf{s}_i$ denotes the size vector (i.e., width and height) of the bounding box $\{\mathbf{x}_{ui}, \mathbf{x}_{di}\}$. $\mathbf{t}^j_{ui}$ and $\mathbf{t}^j_{di}$ are two vectors randomly sampled from a Gaussian distribution. $\odot$ denotes the element-wise multiplication. $\theta$ is a pre-defined scaling factor which can be utilized to control the sample range of the bounding boxes. Benefiting from the random sampling, these bounding boxes can provide a robust approximation to the output of the teacher detector regardless of its generalization performance. Then, we establish two separate supervised learning schemes for both pseudo category polishing network and pseudo bounding-box polishing network using these sampled boxes, namely pseudo category polishing learning and pseudo bounding-box polishing learning. Due to their similar structures, we term these two learning schemes together as dual polishing learning in this study.

\subsubsection{\textbf{Pseudo Category Polishing Learning}}
As mentioned in Section~\ref{sec::classification_polishing_network}, the pseudo category polishing network aims to re-predict the category of the unannotated object based on the bounding box produced by the teacher detector. %In order to improve the classification accuracy of unannotated objects with pseudo bounding boxes close to the ground truth
To this end, we introduce a small $\theta$, termed $\theta_{cls_c}$, to sample $N_{cls_c}$ bounding boxes for each annotated object, then select these overlapped with the ground truth by a large enough IOU (e.g., $\textgreater \tau_{pos}$) as the input of the pseudo category polishing network and encourage the network to predict their categories. We term these bounding boxes as positive samples for the ground truth category. On the other hand, we introduce a large $\theta$, termed $\theta_{cls_m}$, to sample $N_{cls_m}$ bounding boxes and select those overlapped with the ground truth by a small IOU (e.g., $\textless \tau_{pos}$) as the negative samples for the ground truth category. To increase the diversity of negative samples, we also introduce the proposal bounding boxes that are produced by the RPN module in the teacher detector and overlapped with the ground truth by a small IOU (e.g., $\textless \tau_{pos}$)  as negative samples. With all these positive and negative samples, the pseudo category polishing network can be trained in a conventional supervised classification manner.

\subsubsection{\textbf{Pseudo Bounding-Box Polishing Learning}} In order to empower the pseudo bounding-box polishing network to refine the pseudo bounding boxes produced by the teacher detector, we set $\theta$ to a specific value, termed $\theta_{reg}$ and sample $N_{reg}$ bounding boxes for each annotated object as the input of the polishing network and encourage the network to regress the ground truth bounding boxes. By doing this, we can train the pseudo bounding-box polishing network in a conventional supervised regression manner.

\subsubsection{\textbf{Discussion on Bias in Labeled Subset}}
An important issue is how to guarantee well generalization ability (i.e. smaller bias) for object detector with scarce annotated images. For this purpose, a target much simpler than the object detection task is defined in this study first, which is obtained by reducing the deviation between the ground truth and the pseudo labels produced by the teacher model. 
Then, through sufficiently and randomly simulations for the output from the teacher model, the simulated deviation can be abundant and diverse enough, which thus impedes both polishing networks from biasing towards the labeled subset. 
A comparison regarding to the distribution between the simulated deviation and the real deviation is given in Section~\ref{sec::ablation_study}.

\subsection{SSOD with Disentangled Pseudo Labels}
With the proposed dual polishing learning scheme, we can obtain more accurate pseudo labels for SSOD. To introduce more unannotated objects for SSOD and further enhance the performance, we propose to disentangle the polished pseudo categories and bounding boxes of unannotated objects for SSOD, i.e., the polished pseudo category or bounding box for a given unannotated object is separately utilized for category classification or bounding-box regression training in SSOD. Specifically, given the prediction results (i.e., pseudo labels) of the teacher detector, we first pre-define a threshold $\eta$ and select those unannotated objects with a classification confidence higher than $\eta$ as the candidate unannotated objects for dual pseudo-label polishing learning. Then, all unannotated objects with refined pseudo category labels along with a classification confidence higher than a pre-defined threshold $\tau_{cls}$ will be augmented with those annotated objects for category classification training in SSOD. For bounding boxes regression training, all candidate unannotated objects with refined bounding boxes will be augmented with those annotated objects for SSOD.

\section{Experiments}
\subsection{Datasets}
Following previous works~\cite{sohn2020simple,zhou2021instant}, we evaluate the proposed method on two commonly utilized SSOD benchmark
%  dataset
s including PASCAL VOC~\cite{everingham2010pascal} and MS-COCO~\cite{lin2014microsoft}. 

\noindent\textbf{PASCAL VOC}: We employ images from the training-validation set \texttt{trainval} in PASCAL VOC07 as the labeled data, while images from the training-validation set \texttt{trainval} in PASCAL VOC12 as the unlabeled data. \texttt{trainval} in PASCAL VOC07 contains 5,011 images and the \texttt{trainval} in PASCAL VOC12 contains about 11,540 images. In this way, the ratio of labeled data to the unlabeled one is roughly 1:2. For performance evaluation, we adopt the test set \texttt{test} in PASCAL VOC07 and report the commonly utilized mAP metrics~\cite{zhou2021instant,xu2021end}, i.e., $AP_{50}$ and $AP_{50:95}$ over 20 classes.

\begin{table}[t]  
\centering
\caption{Numerical results of different methods on the PASCAL VOC dataset.} 
\resizebox{!}{0.58in}{
\begin{tabular}{ccccc}
    \hline
    Method & Remark & $AP_{50}$ & $AP_{50:95}$ \\
    \hline
    Supervised (Ours) &  & $76.70$ & $43.00$ \\
    % Supervised & VOC0712 & None & $84.10$ & $51.70$ \\
    \hline
    STAC~\cite{sohn2020simple} & arxiv 2020 & $77.45$ & $44.64$ \\
    Unbiased Teacher~\cite{liu2021unbiased} & ICLR 2021 & $77.37$ & $48.69$ \\
    Instant-Teaching~\cite{zhou2021instant} & CVPR 2021 & $78.30$ & $48.70$ \\
    Instant-Teaching$^*$~\cite{zhou2021instant} & CVPR 2021 & $79.20$ & $50.00$ \\
    Humble Teacher~\cite{tang2021humble} & CVPR 2021 & $80.94$ & $53.04$ \\
    \hline
    Ours &  & $82.50$ & $52.40$ \\
    Ours$(scale jitter)$ &  & $\textbf{84.90}$ & $\textbf{55.50}$ \\
    \hline
\end{tabular}
\label{table:VOC_final}
}
\end{table}

\begin{table*}[h]  
\centering
    \caption{Numerical results %of different results 
    on MS-COCO dataset with different amounts of labeled images.} 
    \begin{tabular}{cccccc}
        \hline
        Method & Remark & $1\%$ & $5\%$ & $10\%$ \\
        \hline
        Supervised & & 10.0$\pm$0.26 & 20.92$\pm$0.15 & 26.94$\pm$0.111 \\
        \hline
        STAC~\cite{sohn2020simple} & arxiv 2020 & 13.97$\pm$0.35 & 24.38$\pm$0.12 & 28.64$\pm$0.21  \\
        Unbiased Teacher~\cite{liu2021unbiased} & ICLR 2021 & 20.75$\pm$0.12 & 28.27$\pm$0.11 & 31.50$\pm$0.10 \\
        Instant-Teaching~\cite{zhou2021instant} & CVPR 2021& 16.00$\pm{0.20}$ & 25.50$\pm{0.05}$ & 29.45$\pm{0.15}$ \\
        Instant-Teaching$^*$~\cite{zhou2021instant} & CVPR 2021 & 18.05$\pm{0.15}$ & 26.75$\pm{0.05}$ & 30.40$\pm{0.05}$ \\
        Humble Teacher~\cite{tang2021humble} & CVPR 2021& 16.96$\pm$0.38 & 27.70$\pm$0.15 & 31.61$\pm$0.28 \\
        Soft Teacher~\cite{xu2021end} & ICCV 2021 & 20.46$\pm{0.39}$ & 30.74$\pm{0.08}$  & 34.04$\pm{0.14}$ \\
        Ours & & \textbf{23.55}$\pm{\textbf{0.25}}$ & \textbf{32.10}$\pm{\textbf{0.15}}$ & \textbf{35.30}$\pm{\textbf{0.15}}$ \\
        \hline
    \end{tabular}
    \label{table:COCO_final}
\end{table*}

\noindent\textbf{MS-COCO}: Similar as~\cite{sohn2020simple,zhou2021instant}, we separately randomly select $1\%$, $5\%$ and $10\%$ images from the COCO training set \texttt{train2017} as the labeled data while the remaining are used as unlabeled data to evaluate the SSOD performance under different amounts of labeled data. The training set totally contains 118K images. For test, we evaluate the proposed method on the COCO validation set \texttt{val2017} which consists of 5K images. On this dataset, we take the mAP metric $AP_{50:95}$~\cite{zhou2021instant,xu2021end} for SSOD performance evaluation.

\subsection{Comparison Methods}
In the following experiments, we compare the proposed method with a supervised baseline and five other state-of-the-art SSOD methods, including STAC~\cite{sohn2020simple}, Unbiased Teacher~\cite{liu2021unbiased}, Instant-Teaching~\cite{zhou2021instant}, Humble Teacher~\cite{tang2021humble} and Soft Teacher~\cite{xu2021end}. Among them, STAC~\cite{sohn2020simple} provides a seminal framework for SSOD. Unbiased Teacher~\cite{liu2021unbiased} jointly trains a student and a gradually progressing teacher in a mutually-beneficial manner. Humble Teacher~\cite{tang2021humble} employs the EMA strategy to update the teacher detector from the student one online. Instant-Teaching~\cite{zhou2021instant} and Soft Teacher~\cite{xu2021end} investigate end-to-end SSOD frameworks. In this study, the proposed method also adopts the end-to-end SSOD framework but mainly focuses on reducing the deviation between pseudo labels and ground truth using dual polishing learning. For fair comparison, we exactly follow the same experimental settings on both datasets as these methods and directly report the results released in their provenance in the following tables.

\subsection{Implementation Details}
We implement the proposed dual pseudo-label polishing framework based on the end-to-end SSOD framework~\cite{xu2021end} with removing the soft teacher and box jitter schemes. Following~\cite{liu2021unbiased,xu2021end}, we employ the Faster R-CNN~\cite{NIPS2015_14bfa6bb} with FPN~\cite{Lin_2017_CVPR} as the teacher and student detectors, where the ResNet-50~\cite{he2016deep} with initialized weights pre-trained on ImageNet~\cite{deng2009imagenet} is utilized as the feature representation backbone. The hyper-parameters in detectors are determined according to the MMDetection toolbox~\cite{chen2019mmdetection}. Due to limited space, the SSOD training details as well as the hyper-parameters (e.g., $\theta_{cls_c}$, $N_{cls_c}$, $\theta_{cls_m}$, $N_{cls_m}$, $\tau_{pos}$ etc.) setting for the proposed method are given in the supplementary material. %

\subsection{Performance on Two Benchmarks}
In this part, we compare the proposed method with several state-of-the-art (SOTA) methods on the PASCAL VOC and MS-COCO benchmarks for SSOD. The numerical results can be found in Table~\ref{table:VOC_final} and Table~\ref{table:COCO_final}. As can be seen, on the PASCAL VOC dataset, the proposed method surpasses other methods in most cases. Compared with the supervised baseline, the proposed method improves the $AP_{50}$ and $AP_{50:95}$ by $5\%$ and $9.4\%$, respectively. Even compared with the recently SOTA Humble Teacher~\cite{tang2021humble}, the proposed method improves the $AP_{50}$ by $1.6\%$. %However, $AP_{50:95}$ of the proposed method is inferior to Humble Teacher~\cite{tang2021humble}. This is because the Humble Teacher~\cite{tang2021humble} introduces data augmentation on unannotated images for performance enhancement.
Moreover, it has shown that scale jitter~\cite{xu2021end} %in both strong and weak augmentation 
is beneficial to improve the generalization performance of SSOD. Inspired by this, we also implement a variant of the proposed method by introducing scale jitter for data augmentation. As can be seen, with such a strategy, the performance of the proposed method can be further improved.

On COCO dataset, we report the result of all methods under three different SSOD settings, e.g., with $1\%$, $5\%$, $10\%$ labeled images. As can be seen, the SSOD tasks on this dataset is more challenging than that on the PASCAL VOC dataset. Nevertheless, the proposed method surpasses other methods with clear margins in all cases, especially when the labeled images is few, e.g. $1\%$ labeled images. This demonstrate that the proposed method is still superior over other methods even with more challenging SSOD tasks.

\begin{figure}[htbp]
    \centering
    \subfigure[]{\includegraphics[width=1.0in, height=1.0in]{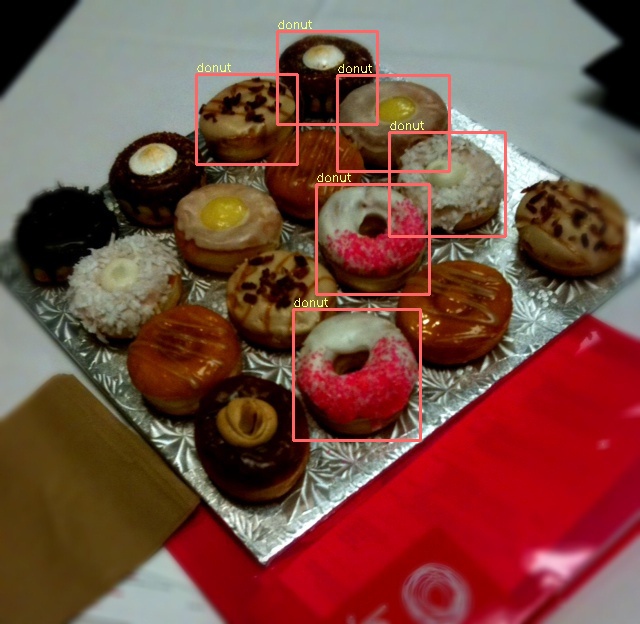}}
    % \quad
    \subfigure[]{\includegraphics[width=1.0in, height=1.0in]{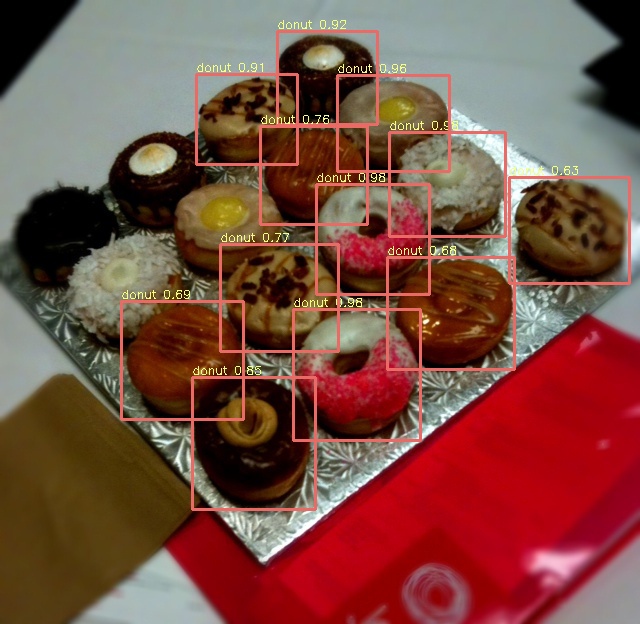}}
    % \quad
    \subfigure[]{\includegraphics[height=1.0in]{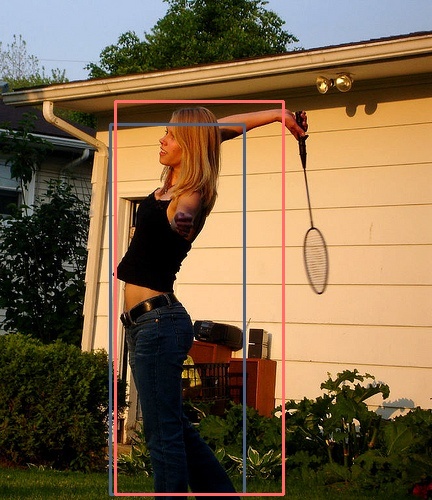}}
    \caption{Visualization result of unannotated objects with polished pseudo labels. Subfigure (a) and (b) illustrate the unannotated objects with pseudo categories (e.g., classification confidence $\textgreater$ 0.9) produced by the teacher detector and polished by the proposed dual polishing learning scheme. (c) illustrate the bounding boxes produced by the teacher detector (e.g., blue box) and polished by the proposed scheme (e.g., red box). As can be seen, the proposed scheme can introduce more unannotated objects with high-quality pseudo labels for SSOD. Best view in color.}
    \label{fig:visual-donut}
\end{figure}

\subsection{Ablation Study}
\label{sec::ablation_study}
In this part, we conduct experiments on the PASCAL VOC dataset to demonstrate the effectiveness of the proposed dual polishing learning, the loss utilized for bounding box regression and the strategy of disentangling pseudo labels, as well as analyse the sensitivity of some key hyper-parameters. 
% More ablation studies are listed in supplementary material.

\begin{table}[t]  
\centering
\caption{Effect of the proposed dual polishing learning on PASCAL VOC dataset.}\label{table:abstudy}
\resizebox{!}{0.36in}{
\begin{tabular}{c|c|cc}
    \hline
    Bounding-box Polishing & Category Polishing & $AP_{50}$ & $AP_{50:95}$ \\
    \hline
    $\times$ & $\times$ & 80.00 & 48.70 \\
    \checkmark & $\times$ & 82.20 & 52.10\\
    $\times$& \checkmark & 82.40 & 51.40 \\
    \checkmark & \checkmark & \textbf{82.50} & \textbf{52.40}\\
    \hline
\end{tabular}
}
\end{table}

\subsubsection{\textbf{Effects of Dual Polishing Learning}}
The key of the proposed method lies on the dual polishing learning which consists of a pseudo category polishing learning scheme and a pseudo bounding-box polishing learning scheme. Thus, it is necessary to conduct ablation study to demonstrate the effectiveness of each one. To this end, we report the detection performance (e.g., $AP_{50}$ and $AP_{50:95}$) of the proposed method with other three variants in Table~\ref{table:abstudy}, which remove the pseudo category polishing learning scheme, the pseudo bounding-box polishing learning scheme or both of them, respectively. As can be seen, when removing either the polishing learning scheme, the detection performance of the proposed method declines. When both of them are removed, the proposed method degenerates to its end-to-end SSOD baseline and the performance declines the most. The effectiveness of the dual polishing learning scheme can be also clarified by the result illustrated in Figure~\ref{fig:coco_reg_polish}, where we can find the accuracy of the pseudo labels can be obviously improved after polishing learning. This can be further clarified by the visualization results in Figure~\ref{fig:visual-donut}. All these results demonstrate that the proposed dual polishing learning is effective for SSOD, especially when coping with challenging dataset.

\begin{table}[htbp]  
\caption{Comparison of the simulated distributions.}\label{table:distributions}
\centering
\subtable[Category (IOU distribution)]{
\centering
\resizebox{!}{0.22in}{
\begin{tabular}{ccc}
    \hline
     & IOU $\ge 0.5$ & IOU $\textless 0.5$ \\
    \hline
    real & mu=0.82, std=0.11 & mu=0.17, std=0.17 \\
    simulated & mu=0.75, std=0.12 & mu=0.07, std=0.14 \\
    \hline
\end{tabular}
\label{table:distri_cate}
}
}
\subtable[Bounding boxes (distribution of position deviation)]{
    \centering
    \resizebox{3.3in}{0.22in}{
    \begin{tabular}{ccccccccc}
    \hline
     & \multicolumn{2}{c}{x1} & \multicolumn{2}{c}{y1} & \multicolumn{2}{c}{x2} & \multicolumn{2}{c}{y2} \\
    \hline
    & mu & std & mu & std & mu & std & mu & std \\
    \hline
    real & 0.0063 & 0.1962 & -0.0087 & 0.1884 & 0.0012 & 0.1968 & 0.0170 & 0.2180 \\
    simulated & -0.0016 & 0.1996 & -0.0011 & 0.1992 & 0.0017 & 0.1996 & 0.0011 & 0.1995 \\
    \hline
\end{tabular}
\label{table:distri_posi}
}
}
\end{table}

\subsubsection{\textbf{Comparison of the Distributions}}
To illustrate the effectiveness of simulated deviation, we experimentally compute the distributions of simulated and real deviations for comparison on COCO dataset ($10\%$), as shown in Table~\ref{table:distributions}. 
As can be seen, the simulated deviation can well approximate the real ones. 
% Experiments also show the well ability of improving the quality of pseudo labels for unannotated objects. 
More analysis can be found in supplementary material.

\begin{table}[htbp]  
\centering
\caption{Different loss functions for the polishing network and the effect of disentangling pseudo labels.}
\begin{tabular}{ccccc} 
    \hline
    Loss & disentanglement &$AP_{50}$ &$AP_{50:95}$\\
    \hline
    $\ell_1$ Loss & $\checkmark$ & 82.10 & 51.80 \\ 
    GIoU Loss & $\times$ & 82.30 & 52.20 \\
    GIoU Loss & $\checkmark$ & \textbf{82.50} & \textbf{52.40} \\
    \hline
\label{table:reg_loss_n_disent}
\end{tabular}
\end{table}

\subsubsection{\textbf{Effect of the GIoU Loss}}
In this study, we utilize GIOU loss~\cite{Rezatofighi_2019_CVPR} instead of the commonly utilized $\ell_1$ norm based loss to train the pseudo bounding-box polishing network. To demonstrate the effectiveness of the GIoU Loss, we compare the proposed method with a variant on PASCAL VOC, which replaces the GIoU loss for polishing network training with $\ell_1$ norm based loss. Their numerical results are reported in Table~\ref{table:reg_loss_n_disent}. As can be seen, GIoU loss based method achieves better polishing performance.

\subsubsection{\textbf{Effect of Disentangling Pseudo Labels}}
In the proposed method, we disentangle the polished pseudo categories and bounding boxes for SSOD. To demonstrate its effectiveness, we compare the proposed method with a variant that polishes the pseudo bounding box and pseudo category for a specific unannotated object in a consecutive way. Their numerical results on the PASCAL dataset is given in Table~\ref{table:reg_loss_n_disent}. As can be seen, disentangling the polished pseudo labels can improve the detection performance, since it will introduce more unannotated objects for SSOD.

% Specifically, for the pseudo categories, we compute the distribution of IOU between the bounding boxes estimated by the teacher model and the ground truth, as well as that between the simulated bounding boxes and the ground truth. For simplicity, we divide the distribution into two parts, including the distribution of IOU $\ge 0.5$ for positive samples (e.g., belonging to the same category as the ground truth) and the distribution of IOU $\textless 0.5$ for negative samples. Each of them is modeled into a Gaussian distribution. For the pseudo bounding boxes, we compute the distribution of position deviation between the estimated or simulated bounding boxes and the ground truth. We also divide the distribution into four parts, namely the distributions of x1,y1,x2,y2 (i.e., upper-left point [x1,y1], down-right point [x2,y2]), and model each of them as a distribution.

\begin{table}[h!]  
\caption{Effect of different hyperparameters}\label{table:hyper}
\centering
\subtable[Effect of $\theta_{reg}$]{
\centering
\resizebox{!}{0.31in}{
\begin{tabular}{cccc}
    \hline
    $\theta_{reg}$ & mean IoU & $AP_{50}$ & $AP_{50:95}$ \\
    \hline
    0.15 & 0.6387$\pm{0.0012}$ & 82.20 & 52.30 \\
    0.2  & 0.5525$\pm{0.0015}$ & \textbf{82.50} & \textbf{52.40} \\
    0.25 & 0.4795$\pm{0.0016}$ & \textbf{82.50} & 52.10 \\
    \hline
\end{tabular}
\label{table:theta_reg}
}
}
\subtable[Effect of $\gamma$]{
    \centering
    \resizebox{!}{0.31in}{
    \begin{tabular}{ccc}
    \hline
    $\gamma$ & $AP_{50}$ & $AP_{50:95}$ \\
    \hline
    0.02 & 82.40 & 51.70 \\
    0.06 & \textbf{82.50} & \textbf{52.40} \\
    0.14 & 82.30 & 52.20 \\
    \hline
\end{tabular}
\label{table:gamma}
}
}
\subtable[Effect of $\eta$]{
    \centering
    \resizebox{!}{0.31in}{
    \begin{tabular}{ccc}
    \hline
    $\eta$ & $AP_{50}$ & $AP_{50:95}$ \\
    \hline
    0.3 & 82.40 & 52.30 \\
    0.5 & \textbf{82.50} & \textbf{52.40} \\
    0.7 & 82.30 & 52.00 \\
    \hline
    \end{tabular}
    \label{table:eta}
}
}
\end{table}

\subsubsection{\textbf{Analysis of Hyper-Parameter Sensitivity}}
The proposed method involves some key hyper-parameters, including the scaling parameter $\theta$ in Eq.~\eqref{eq:random_sample}, the distance parameter $\gamma$ in pseudo bounding-box polishing network, and the initial classification confidence threshold $\eta$. To demonstrate the effect of these hyper-parameters, we test the proposed method with different settings of each of these hyper-parameters on the PASCAL VOC dataset, as shown in Table~\ref{table:hyper}. More analysis details can be found in supplementary material. 

\section{Conclusions}
In this study, we propose a dual pseudo-label polishing framework which mainly focuses on learning to improve the quality of the pseudo labels, viz., reducing the deviation between pseudo labels and ground truth, for accurate SSOD. Specifically, we first elaborately develop two differently structured polishing networks which aim to refine the pseudo categories and bounding boxes produced by a teacher detector. In particular, the pseudo bounding-box polishing network takes a multiple ROI feature fusion scheme to exploit the context of unannotated objects for pseudo bounding boxes refinement. Then, we present a dual polishing scheme where a Gaussian random strategy is utilized to synthesize paired pseudo labels %produced by the teacher detector 
and the corresponding ground truth 
% of categories and bounding boxes 
for supervisedly training both polishing networks, respectively. By doing this, the polishing networks can obviously improve the quality of the pseudo labels for unannotated objects, and thus improve the performance of SSOD. Moreover, such a scheme can be seamlessly plugged into the existing SSOD framework for joint end-to-end learning. In addition, we propose to disentangle the polished pseudo categories and bounding boxes of unannotated objects for SSOD, which enables introducing more different unannotated objects for further performance enhancement. Experiments on both PASCAL VOC and MS-COCO benchmarks demonstrate the efficacy of the proposed method in coping with SSOD, especially those challenging tasks.

\section*{Acknowledgments}
This work was supported in part by the National Natural Science Foundation of China under Grant 62071387, Grand 62101454, and Grant U19B2037; in part by the Shenzhen Fundamental Research Program under Grant JCYJ20190806160210899; in part by the Fundamental Research Funds for the Central Universities.

\bibliography{aaai23}

\begin{thebibliography}{35}
\providecommand{\natexlab}[1]{#1}

\bibitem[{Arazo et~al.(2020)Arazo, Ortego, Albert, O’Connor, and
  McGuinness}]{arazo2020pseudo}
Arazo, E.; Ortego, D.; Albert, P.; O’Connor, N.~E.; and McGuinness, K. 2020.
\newblock Pseudo-labeling and confirmation bias in deep semi-supervised
  learning.
\newblock In \emph{2020 International Joint Conference on Neural Networks
  (IJCNN)}, 1--8. IEEE.

\bibitem[{Bachman, Alsharif, and Precup(2014)}]{bachman2014learning}
Bachman, P.; Alsharif, O.; and Precup, D. 2014.
\newblock Learning with pseudo-ensembles.
\newblock \emph{Advances in neural information processing systems}, 27.

\bibitem[{Berthelot et~al.(2019{\natexlab{a}})Berthelot, Carlini, Cubuk,
  Kurakin, Sohn, Zhang, and Raffel}]{berthelot2019remixmatch}
Berthelot, D.; Carlini, N.; Cubuk, E.~D.; Kurakin, A.; Sohn, K.; Zhang, H.; and
  Raffel, C. 2019{\natexlab{a}}.
\newblock Remixmatch: Semi-supervised learning with distribution matching and
  augmentation anchoring.
\newblock In \emph{International Conference on Learning Representations
  (ICLR)}.

\bibitem[{Berthelot et~al.(2019{\natexlab{b}})Berthelot, Carlini, Goodfellow,
  Papernot, Oliver, and Raffel}]{berthelot2019mixmatch}
Berthelot, D.; Carlini, N.; Goodfellow, I.; Papernot, N.; Oliver, A.; and
  Raffel, C.~A. 2019{\natexlab{b}}.
\newblock Mixmatch: A holistic approach to semi-supervised learning.
\newblock \emph{Advances in Neural Information Processing Systems}, 32.

\bibitem[{Bochkovskiy, Wang, and Liao(2020)}]{bochkovskiy2020yolov4}
Bochkovskiy, A.; Wang, C.-Y.; and Liao, H.-Y.~M. 2020.
\newblock Yolov4: Optimal speed and accuracy of object detection.
\newblock \emph{arXiv preprint arXiv:2004.10934}.

\bibitem[{Chen et~al.(2019)Chen, Wang, Pang, Cao, Xiong, Li, Sun, Feng, Liu, Xu
  et~al.}]{chen2019mmdetection}
Chen, K.; Wang, J.; Pang, J.; Cao, Y.; Xiong, Y.; Li, X.; Sun, S.; Feng, W.;
  Liu, Z.; Xu, J.; et~al. 2019.
\newblock {MMDetection}: Open mmlab detection toolbox and benchmark.
\newblock \emph{arXiv preprint arXiv:1906.07155}.

\bibitem[{Cubuk et~al.(2020)Cubuk, Zoph, Shlens, and Le}]{cubuk2020randaugment}
Cubuk, E.~D.; Zoph, B.; Shlens, J.; and Le, Q.~V. 2020.
\newblock Randaugment: Practical automated data augmentation with a reduced
  search space.
\newblock In \emph{Proceedings of the IEEE/CVF Conference on Computer Vision
  and Pattern Recognition Workshops}, 702--703.

\bibitem[{Deng et~al.(2009)Deng, Dong, Socher, Li, Li, and
  Fei-Fei}]{deng2009imagenet}
Deng, J.; Dong, W.; Socher, R.; Li, L.-J.; Li, K.; and Fei-Fei, L. 2009.
\newblock Imagenet: A large-scale hierarchical image database.
\newblock In \emph{2009 IEEE conference on computer vision and pattern
  recognition}, 248--255. Ieee.

\bibitem[{DeVries and Taylor(2017)}]{devries2017improved}
DeVries, T.; and Taylor, G.~W. 2017.
\newblock Improved regularization of convolutional neural networks with cutout.
\newblock \emph{arXiv preprint arXiv:1708.04552}.

\bibitem[{Everingham et~al.(2010)Everingham, Van~Gool, Williams, Winn, and
  Zisserman}]{everingham2010pascal}
Everingham, M.; Van~Gool, L.; Williams, C.~K.; Winn, J.; and Zisserman, A.
  2010.
\newblock The pascal visual object classes (voc) challenge.
\newblock \emph{International journal of computer vision (IJCV)}, 88(2):
  303--338.

\bibitem[{He et~al.(2016)He, Zhang, Ren, and Sun}]{he2016deep}
He, K.; Zhang, X.; Ren, S.; and Sun, J. 2016.
\newblock Deep residual learning for image recognition.
\newblock In \emph{Proceedings of the IEEE conference on computer vision and
  pattern recognition}, 770--778.

\bibitem[{Iscen et~al.(2019)Iscen, Tolias, Avrithis, and Chum}]{iscen2019label}
Iscen, A.; Tolias, G.; Avrithis, Y.; and Chum, O. 2019.
\newblock Label propagation for deep semi-supervised learning.
\newblock In \emph{Proceedings of the IEEE/CVF Conference on Computer Vision
  and Pattern Recognition}, 5070--5079.

\bibitem[{Lee et~al.(2013)}]{lee2013pseudo}
Lee, D.-H.; et~al. 2013.
\newblock Pseudo-label: The simple and efficient semi-supervised learning
  method for deep neural networks.
\newblock In \emph{Workshop on challenges in representation learning, ICML},
  volume~3, 896.

\bibitem[{Lin et~al.(2017{\natexlab{a}})Lin, Dollar, Girshick, He, Hariharan,
  and Belongie}]{Lin_2017_CVPR}
Lin, T.-Y.; Dollar, P.; Girshick, R.; He, K.; Hariharan, B.; and Belongie, S.
  2017{\natexlab{a}}.
\newblock Feature Pyramid Networks for Object Detection.
\newblock In \emph{Proceedings of the IEEE Conference on Computer Vision and
  Pattern Recognition (CVPR)}.

\bibitem[{Lin et~al.(2017{\natexlab{b}})Lin, Goyal, Girshick, He, and
  Dollar}]{Lin_2017_ICCV}
Lin, T.-Y.; Goyal, P.; Girshick, R.; He, K.; and Dollar, P. 2017{\natexlab{b}}.
\newblock Focal Loss for Dense Object Detection.
\newblock In \emph{Proceedings of the IEEE International Conference on Computer
  Vision (ICCV)}.

\bibitem[{Lin et~al.(2014)Lin, Maire, Belongie, Hays, Perona, Ramanan,
  Doll{\'a}r, and Zitnick}]{lin2014microsoft}
Lin, T.-Y.; Maire, M.; Belongie, S.; Hays, J.; Perona, P.; Ramanan, D.;
  Doll{\'a}r, P.; and Zitnick, C.~L. 2014.
\newblock Microsoft coco: Common objects in context.
\newblock In \emph{European conference on computer vision (ECCV)}, 740--755.
  Springer.

\bibitem[{Liu et~al.(2021)Liu, Ma, He, Kuo, Chen, Zhang, Wu, Kira, and
  Vajda}]{liu2021unbiased}
Liu, Y.-C.; Ma, C.-Y.; He, Z.; Kuo, C.-W.; Chen, K.; Zhang, P.; Wu, B.; Kira,
  Z.; and Vajda, P. 2021.
\newblock Unbiased Teacher for Semi-Supervised Object Detection.
\newblock In \emph{International Conference on Learning Representations}.

\bibitem[{Miyato et~al.(2018)Miyato, Maeda, Koyama, and
  Ishii}]{miyato2018virtual}
Miyato, T.; Maeda, S.-i.; Koyama, M.; and Ishii, S. 2018.
\newblock Virtual adversarial training: a regularization method for supervised
  and semi-supervised learning.
\newblock \emph{IEEE Transactions on Pattern Analysis and Machine Intelligence
  (TPAMI)}, 41(8): 1979--1993.

\bibitem[{Redmon and Farhadi(2018)}]{redmon2018yolov3}
Redmon, J.; and Farhadi, A. 2018.
\newblock Yolov3: An incremental improvement.
\newblock \emph{arXiv preprint arXiv:1804.02767}.

\bibitem[{Ren et~al.(2015)Ren, He, Girshick, and Sun}]{NIPS2015_14bfa6bb}
Ren, S.; He, K.; Girshick, R.; and Sun, J. 2015.
\newblock Faster R-CNN: Towards Real-Time Object Detection with Region Proposal
  Networks.
\newblock In Cortes, C.; Lawrence, N.; Lee, D.; Sugiyama, M.; and Garnett, R.,
  eds., \emph{Advances in Neural Information Processing Systems}, volume~28.
  Curran Associates, Inc.

\bibitem[{Rezatofighi et~al.(2019)Rezatofighi, Tsoi, Gwak, Sadeghian, Reid, and
  Savarese}]{Rezatofighi_2019_CVPR}
Rezatofighi, H.; Tsoi, N.; Gwak, J.; Sadeghian, A.; Reid, I.; and Savarese, S.
  2019.
\newblock Generalized Intersection Over Union: A Metric and a Loss for Bounding
  Box Regression.
\newblock In \emph{Proceedings of the IEEE/CVF Conference on Computer Vision
  and Pattern Recognition (CVPR)}.

\bibitem[{Sajjadi, Javanmardi, and Tasdizen(2016)}]{sajjadi2016regularization}
Sajjadi, M.; Javanmardi, M.; and Tasdizen, T. 2016.
\newblock Regularization with stochastic transformations and perturbations for
  deep semi-supervised learning.
\newblock \emph{Advances in neural information processing systems}, 29.

\bibitem[{Sohn et~al.(2020{\natexlab{a}})Sohn, Berthelot, Carlini, Zhang,
  Zhang, Raffel, Cubuk, Kurakin, and Li}]{sohn2020fixmatch}
Sohn, K.; Berthelot, D.; Carlini, N.; Zhang, Z.; Zhang, H.; Raffel, C.~A.;
  Cubuk, E.~D.; Kurakin, A.; and Li, C.-L. 2020{\natexlab{a}}.
\newblock Fixmatch: Simplifying semi-supervised learning with consistency and
  confidence.
\newblock \emph{Advances in Neural Information Processing Systems}, 33:
  596--608.

\bibitem[{Sohn et~al.(2020{\natexlab{b}})Sohn, Zhang, Li, Zhang, Lee, and
  Pfister}]{sohn2020simple}
Sohn, K.; Zhang, Z.; Li, C.-L.; Zhang, H.; Lee, C.-Y.; and Pfister, T.
  2020{\natexlab{b}}.
\newblock A Simple Semi-Supervised Learning Framework for Object Detection.
\newblock \emph{arXiv preprint arXiv:2005.04757}.

\bibitem[{Song, Liu, and Wang(2020)}]{song2020revisiting}
Song, G.; Liu, Y.; and Wang, X. 2020.
\newblock Revisiting the sibling head in object detector.
\newblock In \emph{Proceedings of the IEEE/CVF Conference on Computer Vision
  and Pattern Recognition}, 11563--11572.

\bibitem[{Tang et~al.(2021)Tang, Chen, Luo, and Zhang}]{tang2021humble}
Tang, Y.; Chen, W.; Luo, Y.; and Zhang, Y. 2021.
\newblock Humble teachers teach better students for semi-supervised object
  detection.
\newblock In \emph{Proceedings of the IEEE/CVF Conference on Computer Vision
  and Pattern Recognition}, 3132--3141.

\bibitem[{Tarvainen and Valpola(2017)}]{tarvainen2017mean}
Tarvainen, A.; and Valpola, H. 2017.
\newblock Mean teachers are better role models: Weight-averaged consistency
  targets improve semi-supervised deep learning results.
\newblock \emph{Advances in neural information processing systems}, 30.

\bibitem[{Xie et~al.(2020{\natexlab{a}})Xie, Dai, Hovy, Luong, and
  Le}]{xie2020unsupervised}
Xie, Q.; Dai, Z.; Hovy, E.; Luong, T.; and Le, Q. 2020{\natexlab{a}}.
\newblock Unsupervised data augmentation for consistency training.
\newblock \emph{Advances in Neural Information Processing Systems}, 33:
  6256--6268.

\bibitem[{Xie et~al.(2020{\natexlab{b}})Xie, Luong, Hovy, and Le}]{xie2020self}
Xie, Q.; Luong, M.-T.; Hovy, E.; and Le, Q.~V. 2020{\natexlab{b}}.
\newblock Self-training with noisy student improves imagenet classification.
\newblock In \emph{Proceedings of the IEEE/CVF conference on computer vision
  and pattern recognition}, 10687--10698.

\bibitem[{Xu et~al.(2021)Xu, Zhang, Hu, Wang, Wang, Wei, Bai, and
  Liu}]{xu2021end}
Xu, M.; Zhang, Z.; Hu, H.; Wang, J.; Wang, L.; Wei, F.; Bai, X.; and Liu, Z.
  2021.
\newblock End-to-end semi-supervised object detection with soft teacher.
\newblock In \emph{Proceedings of the IEEE/CVF International Conference on
  Computer Vision}, 3060--3069.

\bibitem[{Zhai et~al.(2019)Zhai, Oliver, Kolesnikov, and Beyer}]{zhai2019s4l}
Zhai, X.; Oliver, A.; Kolesnikov, A.; and Beyer, L. 2019.
\newblock S4l: Self-supervised semi-supervised learning.
\newblock In \emph{Proceedings of the IEEE/CVF International Conference on
  Computer Vision}, 1476--1485.

\bibitem[{Zhang et~al.(2018)Zhang, Cisse, Dauphin, and
  Lopez-Paz}]{zhang2018mixup}
Zhang, H.; Cisse, M.; Dauphin, Y.~N.; and Lopez-Paz, D. 2018.
\newblock mixup: Beyond Empirical Risk Minimization.
\newblock In \emph{International Conference on Learning Representations}.

\bibitem[{Zhang et~al.(2020)Zhang, Chi, Yao, Lei, and Li}]{zhang2020bridging}
Zhang, S.; Chi, C.; Yao, Y.; Lei, Z.; and Li, S.~Z. 2020.
\newblock Bridging the gap between anchor-based and anchor-free detection via
  adaptive training sample selection.
\newblock In \emph{Proceedings of the IEEE/CVF conference on computer vision
  and pattern recognition}, 9759--9768.

\bibitem[{Zhou et~al.(2021)Zhou, Yu, Wang, Qian, and Li}]{zhou2021instant}
Zhou, Q.; Yu, C.; Wang, Z.; Qian, Q.; and Li, H. 2021.
\newblock Instant-teaching: An end-to-end semi-supervised object detection
  framework.
\newblock In \emph{Proceedings of the IEEE/CVF Conference on Computer Vision
  and Pattern Recognition}, 4081--4090.

\bibitem[{Zoph et~al.(2020)Zoph, Ghiasi, Lin, Cui, Liu, Cubuk, and
  Le}]{zoph2020rethinking}
Zoph, B.; Ghiasi, G.; Lin, T.-Y.; Cui, Y.; Liu, H.; Cubuk, E.~D.; and Le, Q.
  2020.
\newblock Rethinking pre-training and self-training.
\newblock \emph{Advances in neural information processing systems}, 33:
  3833--3845.

\end{thebibliography}

\end{document}